\documentclass[runningheads]{llncs}
\usepackage{graphicx}
\usepackage{subcaption}
\usepackage{amsmath}
\DeclareMathOperator*{\argmin}{\arg\!\min}
\DeclareMathOperator*{\argmax}{\arg\!\max}
\newcommand{\norm}[1]{\left\lVert#1\right\rVert}

\begin{document}
\title{SUPER-IVIM-DC: Intra-voxel incoherent motion based Fetal lung maturity assessment from limited DWI data using supervised learning coupled with data-consistency
\thanks{This research was supported in part by a grant from the United States-Israel Binational Science Foundation (BSF), Jerusalem, Israel.}
}
\titlerunning{Fetal lung maturity assessment from limited DW-MRI data}
%
\author{Noam Korngut\inst{1} \and
Elad Rotman\inst{1} \and
Onur Afacan\inst{2}\orcidID{0000-0003-2112-3205} \and
Sila Kurugol\inst{2}\orcidID{0000-0002-5081-4569} \and
Yael Zaffrani-Reznikov\inst{1} \and
Shira Nemirovsky-Rotman\inst{1}\orcidID{0000-0003-4598-1589} \and
Simon Warfield\inst{2}\orcidID{0000-0002-7659-3880} \and 
Moti Freiman\inst{1}\orcidID{0000-0003-1083-1548}}
\authorrunning{N. Korngut et al.}
%
\institute{Faculty of Biomedical Engineering, Technion, Haifa, Israel \and
Boston Children's Hospital, Boston, MA, USA 
\\
\email{noam.korngut@campus.technion.ac.il}
}
\maketitle              

\begin{sloppypar}

\begin{abstract}
Intra-voxel incoherent motion (IVIM) analysis of fetal lungs Diffusion-Weighted MRI (DWI) data shows potential in providing quantitative imaging bio-markers that reflect, indirectly, diffusion and pseudo-diffusion for non-invasive fetal lung maturation assessment. However, long acquisition times, due to the large number of different ``b-value'' images required for IVIM analysis, precluded clinical feasibility. 
We introduce SUPER-IVIM-DC a deep-neural-networks (DNN) approach which couples supervised loss with a data-consistency term to enable IVIM analysis of DWI data acquired with a limited number of b-values.
We demonstrated the added-value of SUPER-IVIM-DC over both classical and recent DNN approaches for IVIM analysis through numerical simulations, healthy volunteer study, and IVIM analysis of fetal lung maturation from fetal DWI data. 
Our numerical simulations and healthy volunteer study show that SUPER-IVIM-DC estimates of the IVIM model parameters from limited DWI data had lower normalized root mean-squared error compared to previous DNN-based approaches. Further, SUPER-IVIM-DC estimates of the pseudo-diffusion fraction parameter from limited DWI data of fetal lungs correlate better with gestational age compared to both to classical and DNN-based approaches (0.555 vs. 0.463 and 0.310). 
SUPER-IVIM-DC has the potential to reduce the long acquisition times associated with IVIM analysis of DWI data and to provide clinically feasible bio-markers for non-invasive fetal lung maturity assessment.


\end{abstract}
\section{Introduction}
Normal fetal lung parenchyma development starts in the second trimester and progresses through multiple phases before becoming fully functional at full term.  The first phase of development, the embryonic and pseudoglandular stage, is followed by the canalicular phase which starts at 16 weeks, then the saccular stage which starts at 24 weeks, and ends with the alveolar stage which starts at 36 weeks gestation. \cite{schittny2017development}. The progression through these phases is characterized with the formation of a dense capillary network and a progressive increase in pulmonary blood flow, leading to an increased perfusion \cite{Ercolani2021IntraVoxelMaturation}. Maldevelopment of the fetal lung parenchyma, however, may lead to life-threatening physiologic dysfunction due to pulmonary hypoplasia and pulmonary hypertension \cite{lakshminrusimha2015persistent}. The ability to accurately assess lung maturation prior to delivery is therefore critical as newborns with inadequate in-utero lung development are at risk for post-natal respiratory failure or death \cite{ahlfeld2014assessment}. 

In current practice, the non-invasive assessment of fetal lung parenchyma develeopment is delivered by two anatomical imaging modalities:  ultrasonography \cite{Moeglin2005FetalUltrasound}, and magnetic resonance imaging (MRI) \cite{deshmukh2010mr}. Unfortunately, these modalities fail to provide adequate insight into lung function and are therefore suboptimal in assessing fetal lung maturity.

Diffusion-weighted MRI (DWI) is a non-invasive imaging technique sensitive to the random movement of individual water molecules. The displacement of individual water molecules results in signal attenuation in the presence of magnetic field encoding gradient pulses. This attenuation increases with the degree of sensitization-to-diffusion of the MRI pulse sequence (the “b-value”) \cite{Afacan2016FetalAge}. 

It is well established that the DWI signal attenuation essentially encapsulates not just thermally driven water diffusion, but also pseudo-diffusion that results from the randomness of the collective motion of blood water molecules in the randomly oriented network of micro-capillaries \cite{le2019can}. Specifically, the pseudo-diffusion phenomenon is known to attenuate the DWI signal acquired with low b-values (0-200 s/mm$^2$) \cite{freiman2012vivo}.

Several studies demonstrated the potential role of diffusion and pseudo-diffusion imaging biomarkers derived from fetal lung DWI data with the so-called ``intra-voxel incoherent motion'' (IVIM) bi-exponential signal decay model in assessing lung maturity \cite{jakab2018microvascular,Ercolani2021IntraVoxelMaturation}. 
However, long acquisition times required to collect multiple b-values suitable for IVIM-based analysis of fetal lungs diffusion and pseudo-diffusion from DWI data (up to 16 different b-values \cite{jakab2018microvascular}) hinder DWI-based analysis of fetal lung maturation in the clinical setting.

Previously, Bayesian approaches were proposed to address the challenge of obtaining reliable IVIM analyses from low SNR DW-MRI data. Neil et al. \cite{neil1993use} and Orton et al. \cite{orton2014improved} suggested using a Bayesian shrinkage prior. Freiman et al. proposed a spatial homogeneity prior \cite{freiman2013reliable} and recently Spinner et al. \cite{spinner2021bayesian} combined spatial and hierarchical priors. However, their heavy computational burdens hampered their utilization in practice. 

In the past few years, state-of-the art deep-neural-networks (DNN)-based methods were introduced for IVIM parameter estimates. Bertleff et al. \cite{bertleff2017diffusion} demonstrated the ability of supervised DNN to predict the IVIM model parameters from low signal to noise ratio (SNR) DW-MRI data. Barbieri et al. \cite{barbieri2020deep} proposed an unsupervised physics-informed DNN (IVIM-NET) with results comparable to Bayesian methods with further optimizations by Kaandorp et al. \cite{kaandorp2021improved} (IVIM-NET$_{\textrm{optim}}$). Recently, Vasylechko et al. used unsupervised convolutional neural networks (CNN) to improve the reliability of IVIM parameter estimates by leveraging spatial correlations in the data \cite{vasylechko2022self}. Specifically, Zhang et al. used a multi-layer perceptron with an amortized Gaussian posterior to estimate the IVIM model parameters from fetal lung DW-MRI data \cite{zhang2019implicit}.
Yet, long acquisition times due to the large number of different b-value images required for fetal lung IVIM analysis with DNN-based methods (for example 16 in \cite{zhang2019implicit,jakab2018microvascular}) again precluded clinical feasibility. 

In this work, we address the challenge of IVIM parameter estimates from clinical DWI data acquired with a limited number of b-values (i.e. up to 6 b-values) by presenting SUPER-IVIM-DC, a supervised DNN coupled with a data-consistency term that enables the analysis of diffusion and pseudo-diffusion biomarkers. We demonstrated the added value of SUPER-IVIM-DC compared to both the classical least-squares approach and recently proposed DNN-based approaches in assessing diffusion and pseudo-diffusion biomarkers through numerical simulations and a healthy volunteer study. Finally, we demonstrated clinical significance in fetal lung maturity analysis with the pseudo-diffusion fraction parameter of the IVIM model computed from DWI data with a limited number of b-values.

\section{Method}
\subsection{The ``Intra-voxel Incoherent motion'' model of DWI}
The ``Intra-voxel Incoherent motion'' model (IVIM) proposed by Le-Bihan \cite{le2019can}, models the overall MRI signal attenuation as a sum of the diffusion and pseudo-diffusion components taking the shape of a bi-exponential decay:
\begin{equation}
    s_i=s_0 \left(f \exp\left(-b_i \left(D^*+D\right)\right)+\left(1-f\right) \exp\left(-b_i D\right) \right)
    \label{eq:ivim_model}
\end{equation}
where $s_i$ is the signal at b-value $b_i$; $s_0$ is the signal without sensitizing the diffusion gradients; $D$ is the diffusion coefficient, an indirect measure of tissue cellularity; $D^*$ is the pseudo-diffusion coefficient, an indirect measure of blood flow in the micro-capillaries network; and $f$ is the fraction of the contribution of the pseudo-diffusion to the signal decay, which is related to the percentage volume of the micro-capillary network within the voxel. 

Classical methods aim to estimate the IVIM model parameters from the observed data by using either a maximum-likelihood approach:
\begin{equation}
    \hat{\theta} = \argmax_{\theta} L\left(\theta | \left\{s_i\right\}_{i=0}^N\right) = 
                   \argmax_{\theta} P\left(\left\{s_i\right\}_{i=0}^N | \theta \right)
\end{equation}
where $\left\{s_i\right\}_{i=0}^N$ is the set of the observed signals at the different b-values and $\theta =\left\{D,D^*,f\right\}$ are the unknown IVIM model parameters, or a maximum a posterior approach: 
\begin{equation}
    \hat{\theta} = \argmax_{\theta} P\left(\theta|\left\{s_i\right\}_{i=0}^N\right) \propto
    P\left(\left\{s_i\right\}_{i=0}^N|\theta\right)P\left(\theta\right)
\end{equation}
where $P\left(\theta\right)$ is the prior assumed on the distribution of the IVIM model parameters.

In contrast, DNN-based methods formalized the IVIM model parameters estimation problem as a prediction problem:
\begin{equation}
     \hat{\theta} = g_{\Phi}\left(\left\{s_i\right\}_{i=0}^N\right) 
     \label{eq:dnn_predict}
\end{equation}
where $\Phi$ are the DNN weights. The DNN weights are obtained through the minimization of some loss function over a training dataset. Bertleff et al. \cite{bertleff2017diffusion} used a supervised approach to obtain the DNN weights $\Phi$:
\begin{equation}
  \hat{\Phi} = \argmin_\Phi \sum_{k=1}^{K} \norm{ g_{\Phi}\left(\left\{s_i^k\right\}_{i=0}^N\right)-\theta_{ref}^k}^2 
      \label{eq:dnn_train_super}
\end{equation}
where $k\in\{1,...,K\}$ is the index of the DWI signal sample, $\Phi$ are the DNN weights for which the optimization is applied, by using simulated data generated from the reference parameters $\theta_{ref}$ with Eq.~\ref{eq:ivim_model}.
More recently, Barbieri et al. \cite{barbieri2020deep} and Kaandorp et al. \cite{kaandorp2021improved} used IVIM-NET, an unsupervised approach to obtain the DNN weights $\Phi$:
\begin{equation}
      \hat{\Phi} = \argmin_\Phi \sum_{k=1}^{K}\sum_{i=1}^{N} \norm{ f_{g_{\Phi}\left(\left\{s_i^k\right\}_{i=0}^N\right)}\left(s_{0}^{k},b_i\right)-s_{i}^{k}}^2 
      \label{eq:dnn_train_unsupervised}
\end{equation}
where $f_{g_{\Phi}\left(\left\{s_i^k\right\}_{i=0}^N\right)}\left(s_{0}^{k},b_i\right)$ is the signal generated by the IVIM forward model (Eq.~\ref{eq:ivim_model}) given the parameter estimates predicted by the DNN $g_{\Phi}\left(\left\{s_i\right\}_{i=0}^N\right)$ for a given DWI signal $s_{0}^{k}$ and a specific ``b-value'' $b_i,i \in \{1,...,N\}$.

However, these methods may converge to a local minimum during the training process – resulting in suboptimal IVIM predictions when applied to new unseen data. Therefore, DWI data acquired with a large number of b-values are required to sufficiently constrain the DNN  training process, and to obtain accurate IVIM model parameters predictions.

\subsection{SUPER-IVIM-DC}
In a shift from previous approaches, we propose to alleviate the need to acquire DWI data with a large number of ``b-values'' by constraining the DNN training process through a supervised loss function coupled with a data consistency term.  Formally, our training process is defined as: 
\begin{equation}
      \hat{\Phi} = \argmin_\Phi \sum_{k=1}^{K} \left(L_{super}\left(\widehat{\theta^k}, \theta_{ref}^k\right) + \alpha_{dc}L_{dc}\left(\left\{\widehat{s_i^k}\right\}_{i=0}^N, \left\{s_{i,ref}^k\right\}_{i=0}^N\right)\right)
      \label{eq:dnn_train_super_dc}
\end{equation}
where $\widehat{\theta^k}$ are the IVIM model parameters predicted by the DNN for sample $k$, $\theta_{ref}^k$ are the reference parameters used to simulate the DWI data, $\left\{\widehat{s_i^k}\right\}_{i=0}^N$ are the DWI signals at the different ``b-values'' generated with the IVIM forward model (Eq.~\ref{eq:ivim_model}) using the predicted parameters $\widehat{\theta^k}$ and the $\left\{s_{i,ref}^k\right\}_{i=0}^N$ are the corresponding reference DWI signals simulated from the reference parameters $\theta_{ref}^k$.
To accommodate the differences in the IVIM model parameters magnitudes we decomposed $L_{super}\left(\widehat{\theta^k}, \theta_{ref}^k\right)$ into a weighted sum of the squared of the errors in the different IVIM model parameters estimates: 
\begin{equation}
    L_{super}\left(\widehat{\theta^k}, \theta_{ref}^k\right) = \alpha_D\norm{\widehat{D^k}-D^k_{ref}}^2 + \alpha_f\norm{\widehat{f^k}-f^k_{ref}}^2 + \alpha_{D^{*}}\norm{\widehat{{D^{*}}^k}-{D^{*}}^k_{ref}}^2
\end{equation}
in which our data-consistency term is defined as:
\begin{equation}
    L_{dc}\left(\left\{\widehat{s_i^k}\right\}_{i=0}^N, \left\{s_i^k\right\}_{i=0}^N\right) = 
    \sum_{i=1}^{N}\norm{ f_{g_{\Phi}\left(\left\{s_i^k\right\}_{i=0}^N\right)}\left(s_{0}^{k},b_i\right)-s_{i}^{k}}^2 
\end{equation}

\subsection{Implementation details}
We modified the original IVIM-NET$_{\textrm{optim}}$ implementation \cite{kaandorp2021improved} to include our SUPER-IVIM-DC loss function\footnote{Our code and trained models are available on GitHub:\\ https://github.com/TechnionComputationalMRILab/SUPER-IVIM-DC}. The scaling parameters in the SUPER-IVIM-DC loss function ($\{\alpha_{D}, \alpha_{D^*}, \alpha_{f}, \alpha_{dc}\}$)  were considered as hyperparameters. We used a one-dimensional grid search to determine the best values for these hyperparameters. 
We sampled sets of IVIM parameters at relevant ranges for each experiment, and used Eq.~\ref{eq:ivim_model} to simulate DWI signals. To better mimic real data we added Rician Noise, at relevant pre-defined SNR levels, to those datasets. 
We split the simulated data into 90\% for training and 10\% for validation. We implemented our models on Spyder 4.2.0., Python 3.8.5 with PyTorch 1.6.0. We used an Adam optimizer \cite{kingma2014adam} with learning-rate of $10^{-4}$, and a batch-size of 128 to train the DNN. After 10 epochs with no improvement on the validation loss the networks stopped the training and the best performing model was saved for the evaluation. 

\subsection{Evaluation methodology}
\subsubsection{Numerical simulations} 
We conducted a simulation study to analyze the estimation errors for using a various undersampled b-values.
First, we generated one million simulated signals using Eq.~\ref{eq:ivim_model} with various undersampled b-values from the following low b-values: (0,15,30,45,60,75,90,105,120,135,150,175) sec/mm$^2$ concatenated with four high b-values: (200,400,600,800) sec/mm$^2$ and pseudo-random values of Dt, Fp, and Dp.
We sampled every $k^{th}$ b-value from the low b-values vector according to a sampling factor $k\in\{1,...,6\}$, where when $k=1$ represents a full sampling with all b-values. The b-values 0, 200 sec/mm$^2$ remain part of the sampled b-values vector.
We added noise to the simulated signals such that the SNR will be 10. 
Then, we trained SUPER-IVIM-DC and the baseline method IVIM-NET$_{\textrm{optim}}$ using the training data. 

We then simulated 1000 samples of test data using the same approach as we used for training data generation. We estimated the model parameters from the noisy DW-MRI data as a function of the b-values sampling factor. We compared the estimation error of SUPER-IVIM-DC by means of normalized root-mean-squared error (NRMSE) in comparison to the baseline method IVIM-NET$_{\textrm{optim}}$.

\subsubsection{Healthy volunteer study}
\begin{figure}[t]
\centering
\includegraphics[width=1\textwidth]{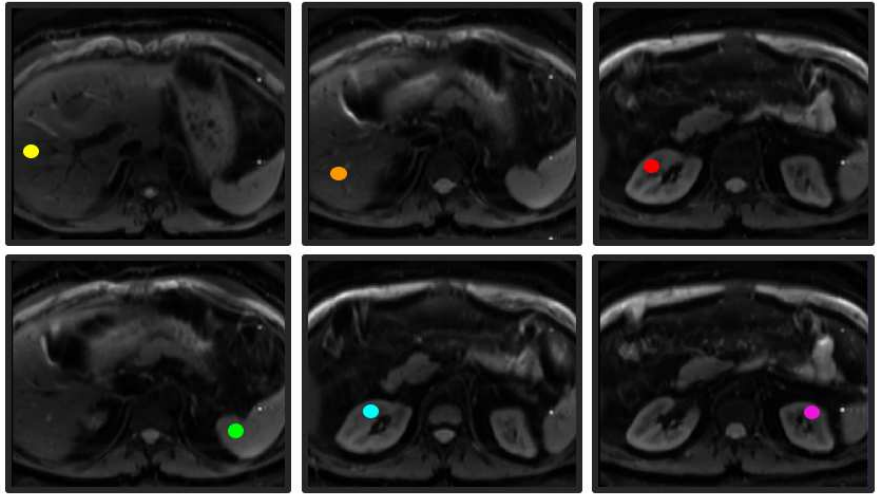} 
    \caption {Abdominal ROI from different slices}
\label{fig:abdominal}
\end{figure}

We assessed the added-value of SUPER-IVIM-DC in estimating the IVIM model parameter values from limited DWI data in a clinical setting. We first acquire high-quality DWI data of a human volunteer with 22 b-values: (0, 12.5, 25, 37.5, 50, 62.5, 75, 87.5, 100, 112.5, 125, 150, 175, 200, 225, 250, 375, 500, 625, 750, 875, 1000) sec/mm$^2$ using a  multi-slice single shot echo planar imaging (EPI) sequence with 6 directions.
We calculated trace-weighted images from the different directions at each b-value using geometric averaging. We annotated 6 different regions of interest in the kidneys, liver and spleen selected from different slices. (Fig.~\ref{fig:abdominal}). 

We estimated reference IVIM model parameters from the entire DWI data using standard least-squares approach.
Next, we sampled the low b-value images (b-value $\leq$ 200 sec/mm$^2$) in the DWI data in a similar way as in our numerical simulations above. 
We trained SUPER-IVIM-DC and the baseline method IVIM-NET$_{\textrm{optim}}$ using simulated data as described above and use them to predict IVIM model parameters from the human volunteer sampled data.

We compared the the estimation error of SUPER-IVIM-DC for the 6 ROI by means of NRMSE in comparison to the baseline method IVIM-NET$_{\textrm{optim}}$ with the high-quality estimates as the reference.

\subsubsection{Clinical impact - correlation between fetal lung maturation and pseudo-diffusion fraction}
We used a legacy fetal lungs DWI data with reference gestational age from a study by Afacan et al. \cite{Afacan2016FetalAge}. 
We used 38 cases of DWI data consisted of 6 b-values:  (0, 50, 100, 200, 400, 600 s/mm$^2$). Detailed description of the acquisition protocol is provided in Afacan et al. \cite{Afacan2016FetalAge}.
We trained SUPER-IVIM-DC and the baseline method IVIM-NET$_{\textrm{optim}}$ using simulated data as described above with the same b-values used in the acquisition and SNR of 10.
We estimated the pseudo-diffusion fraction parameter of the IVIM model ($f$) using classical least-squares approach \cite{gurney2018comparison}, the IVIM-NET$_{\textrm{optim}}$ approach and SUPER-IVIM-DC. 

We assessed Pearson-correlation between the estimated pseudo-diffusion fraction parameter of the IVIM model ($f$) obtained with the different approaches and the gestational age (GA) for different developmental stages.

\section{Results}
\subsubsection{Numerical simulations}
\begin{figure}[t]
\centering
\begin{subfigure}{.32\textwidth}
    \includegraphics[width=\textwidth]{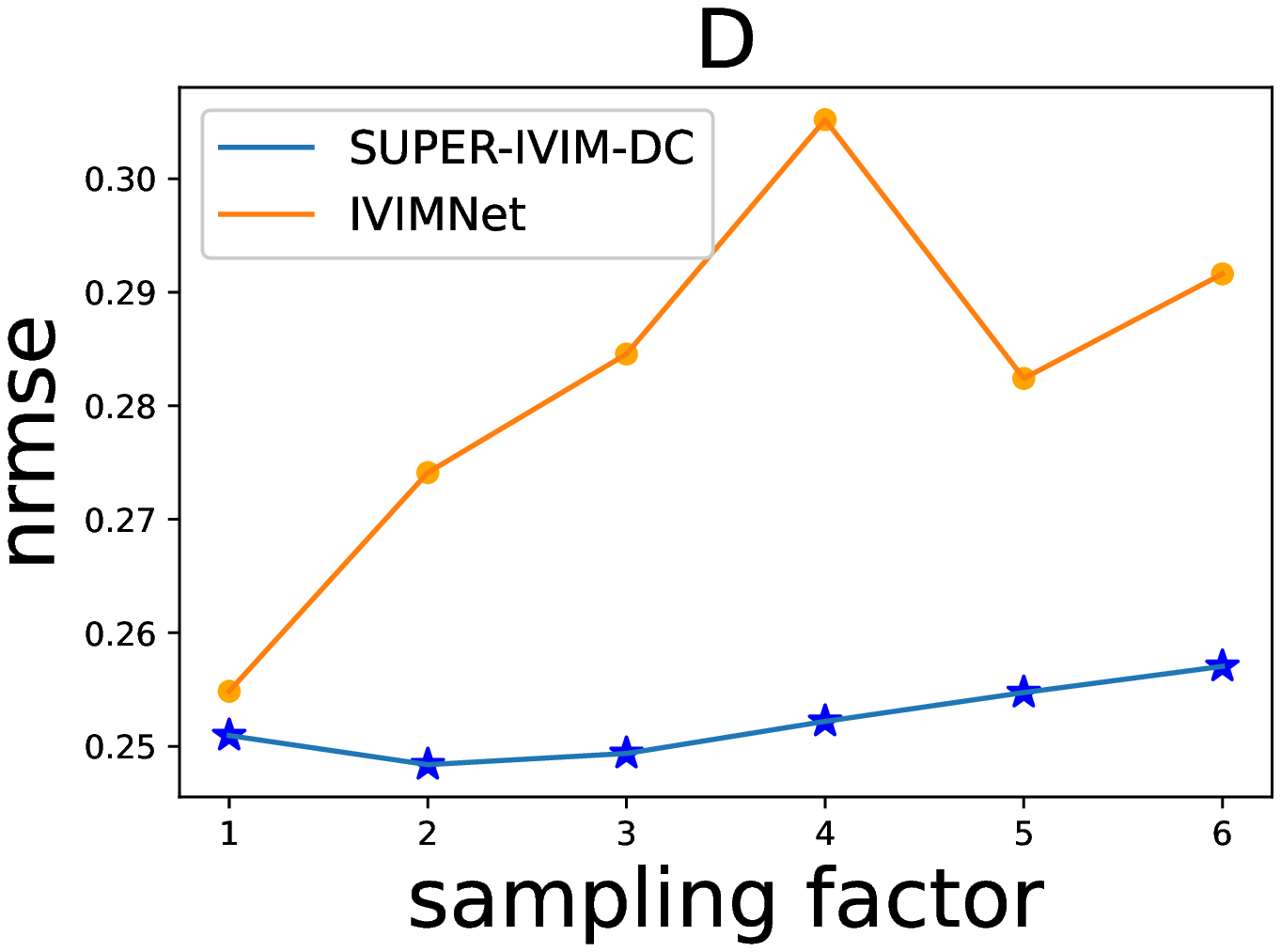} 
    \caption {}
\end{subfigure}
\begin{subfigure}{.32\textwidth}
   \includegraphics[width=\textwidth]{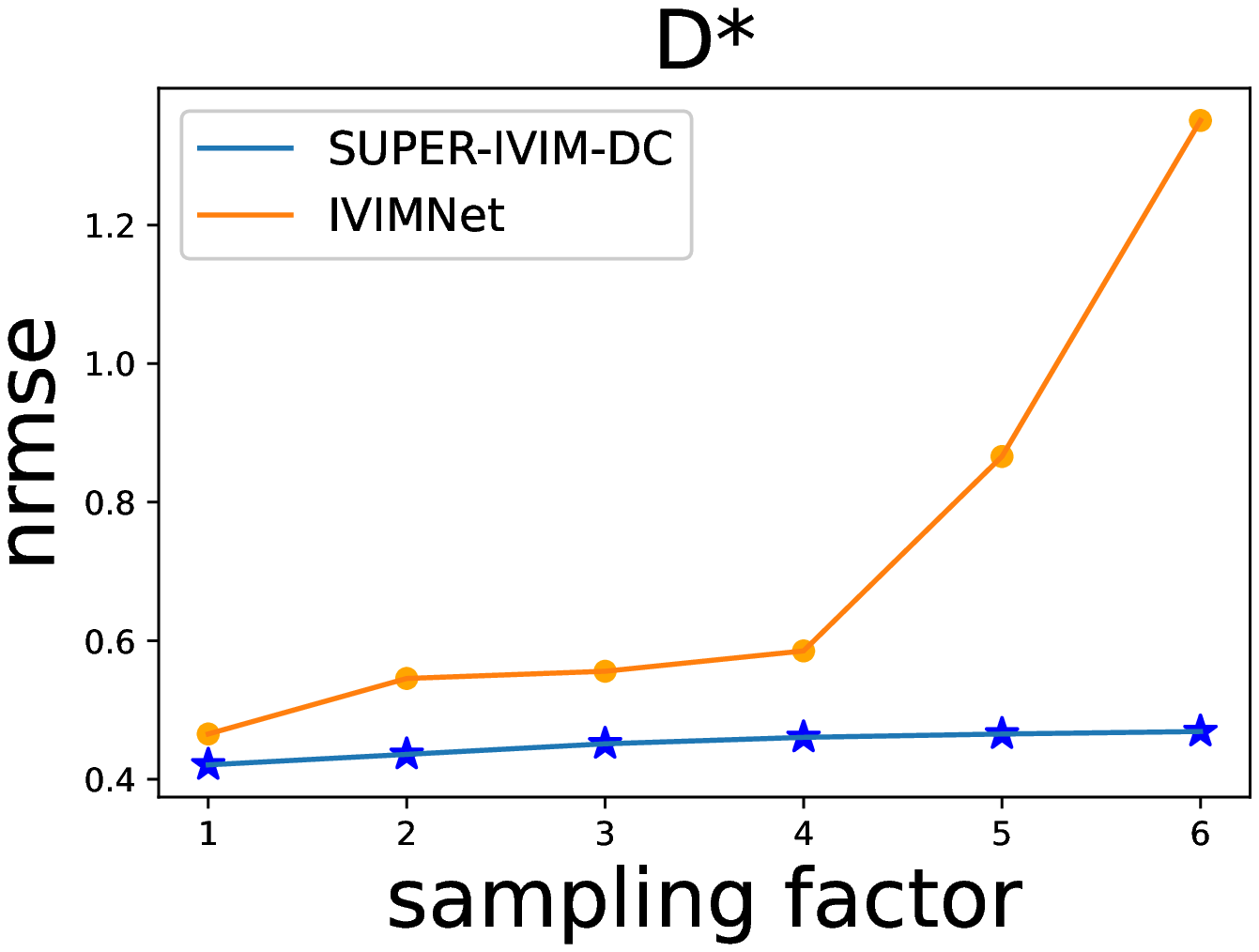} 
    \caption {}
\end{subfigure}
\begin{subfigure}{.32\textwidth}
   \includegraphics[width=\textwidth]{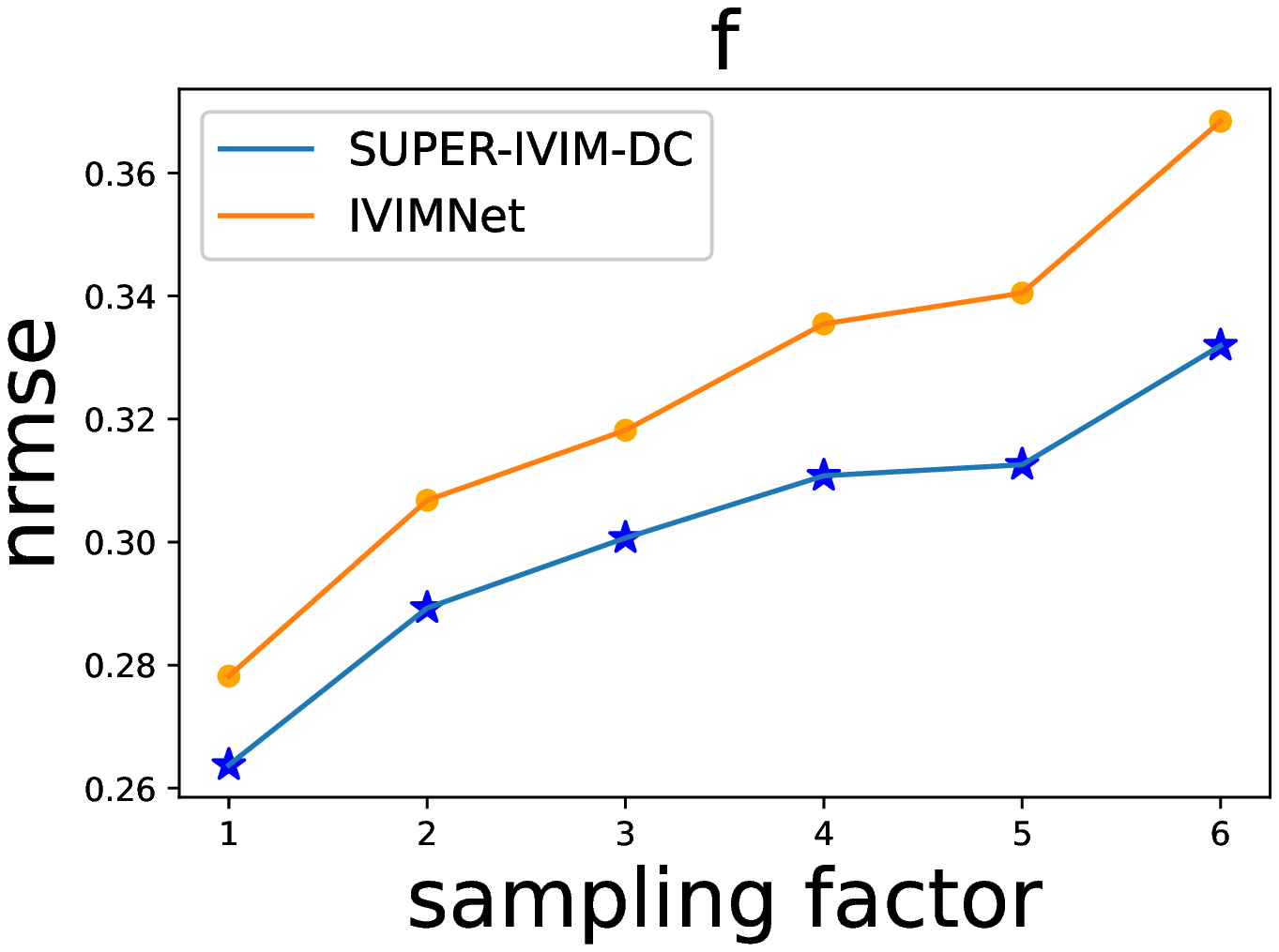} 
    \caption {}
\end{subfigure}
\caption{Numerical simulations results. The NRMSE of the IVIM model parameter estimates obtained with SUPER-IVIM-DC and IVIM-NET$_{optim}$ as a function of the sampling factor.}
\label{fig:numerical_simulations}
\end{figure}

Fig.~\ref{fig:numerical_simulations} summarizes our numerical simulations results. SUPER-IVIM-DC had a lower NRMSE for all IVIM model parameter estimates in all sampling factors compared to IVIM-NET$_{\textrm{optim}}$. The reduced NRMSE obtained with SUPER-IVIM-DC suggests a better generalization of the IVIM model by SUPER-IVIM-DC compared to IVIM-NET$_{\textrm{optim}}$.  

\subsubsection{Healthy volunteer study}
\begin{figure}[t]
\centering
\begin{subfigure}{.32\textwidth}
    \includegraphics[width=\textwidth]{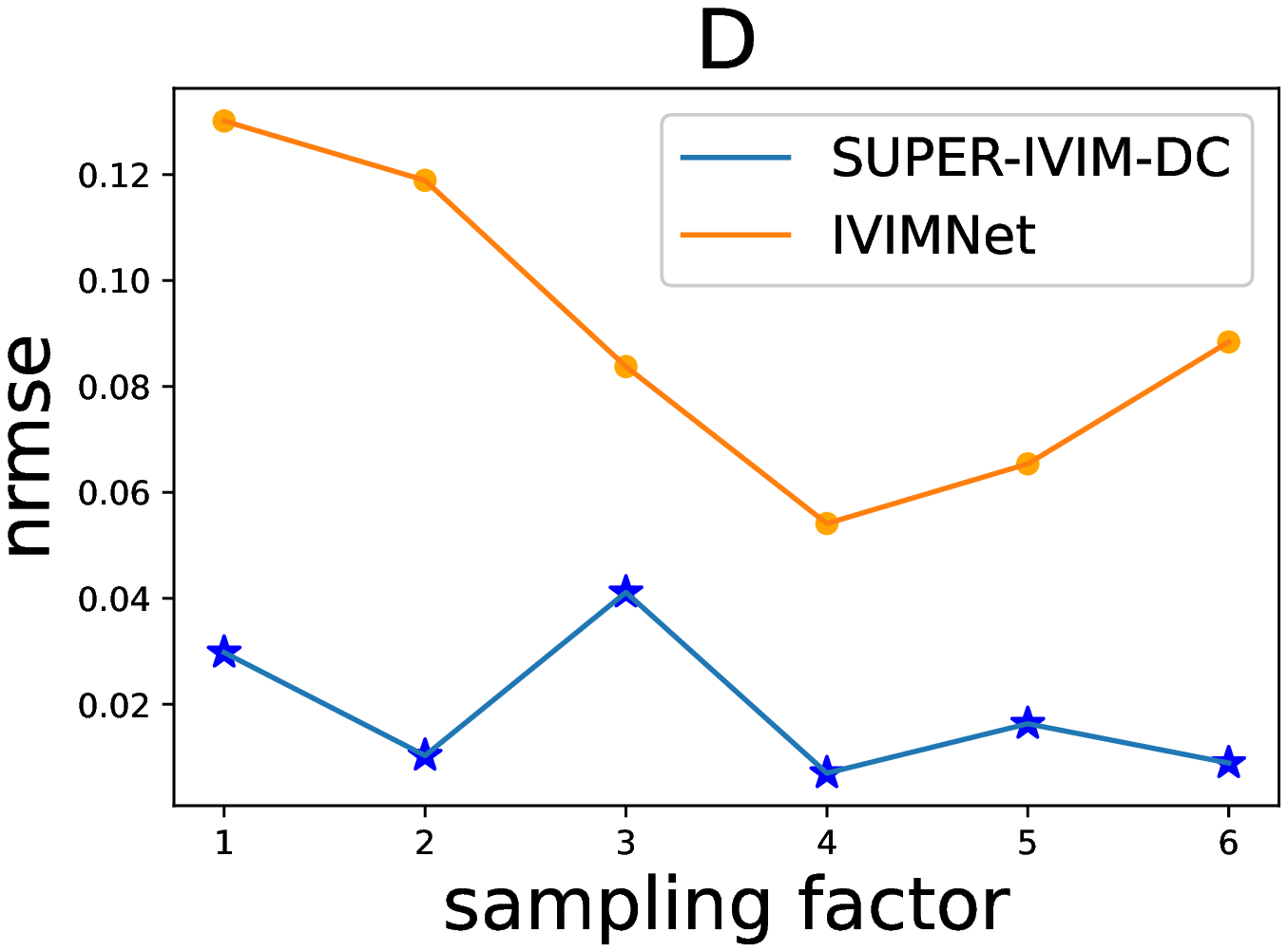} 
    \caption {$D$}
\end{subfigure}
\begin{subfigure}{.32\textwidth}
    \includegraphics[width=\textwidth]{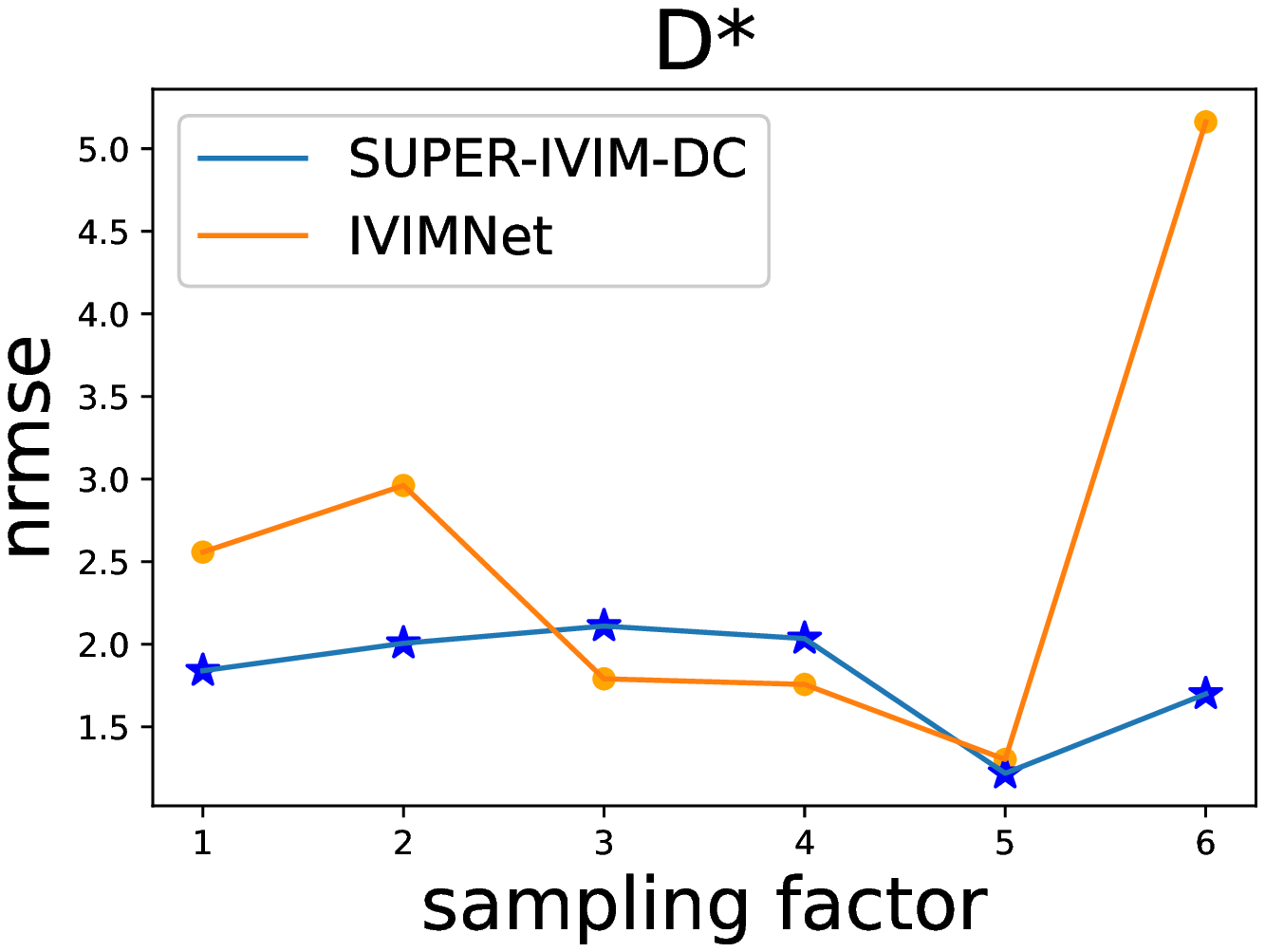} 
    \caption {$D^*$}
\end{subfigure}
\begin{subfigure}{.32\textwidth}
    \includegraphics[width=\textwidth]{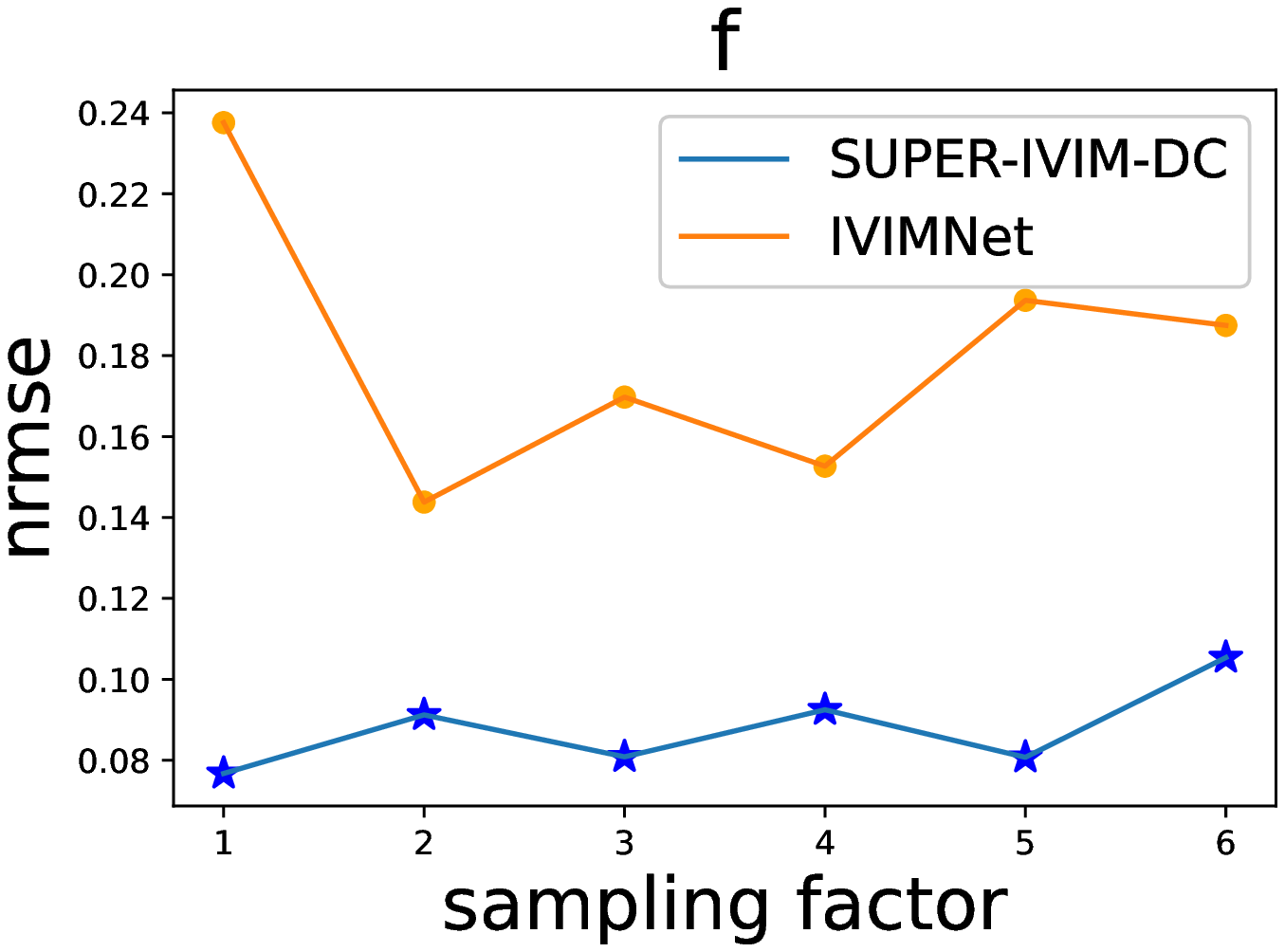} 
    \caption {$f$}
\end{subfigure}
\caption{Healthy volunteer study. SUPER-IVIM-DC reduced IVIM model parameter estimates NRMSE for $D,f$ for all sampling factors. For $D^*$, the reduction was evident only in in sampling factors 1,2, and 6.}
\label{fig:healthy}
\end{figure}

Fig.~\ref{fig:healthy} summarizes our healthy volunteer study results. SUPER-IVIM-DC achieved lower NRMSE compared to IVIM-NET$_{\textrm{optim}}$ for both the $D$ and the $f$ parameters of the IVIM model in all sampling factors. For $D^*$, the improvement was evident in sampling factors 1,2, and 6, while in sampling factors 3-5 the difference was negligible. 

The reduction in NRMSE achieved by SUPER IVIM DC was statistically significant ($p \ll 0.01$) for D and f parameters in experiments 1 and 2. The difference in D* was not statistically significant.

\subsubsection{Clinical impact - correlation between fetal lung maturation and pseudo-diffusion fraction}

\begin{figure}[t]
\includegraphics[width=\textwidth]{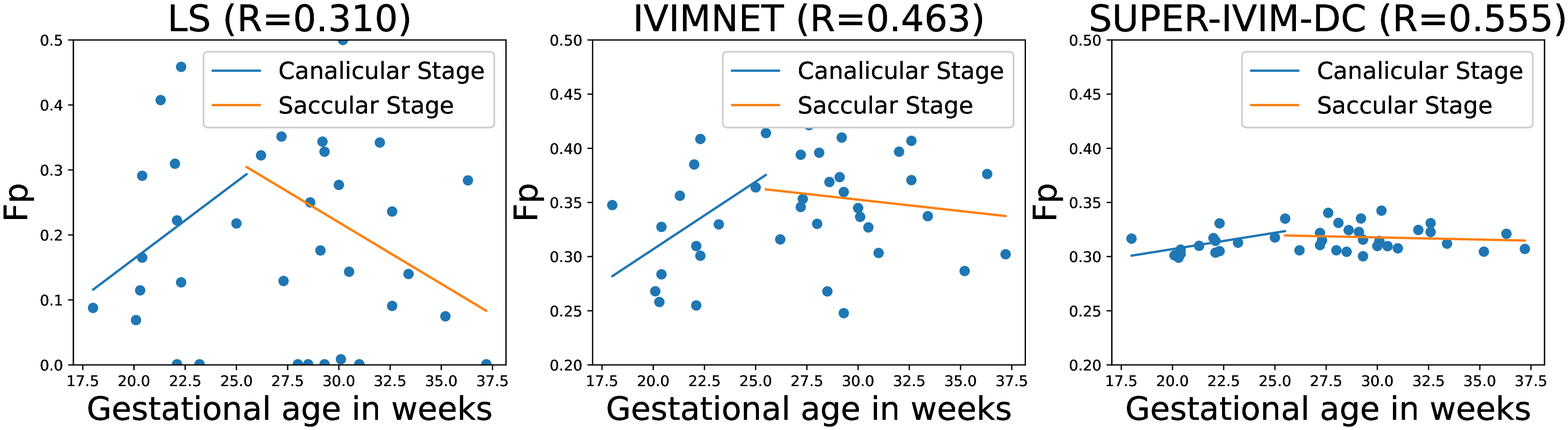}  
\caption{Correlation between the pseudo-diffusion-fraction parameter ($f$) estimated with the different approaches and the gestational age (GA).}
\label{fig:fetal}
\end{figure}

Fig.~\ref{fig:fetal} depicts the correlations between the pseudo-diffusion-fraction of the IVIM model $f$ as estimated by the different methods and the reference GA. By dividing the GA axis into two stages of the fetal lung development - the Canalicular phase (weeks 16-25) and the Saccular phase (week 26-34) our approach demonstrated a better correlation between the $f$ parameter and the GA (0.555 with our SUPER-IVIM-DC vs. 0.463 and 0.310  with IVIMNET and LS) for the Canalicular phase of development (note that LS is in a different scale due to large distribution). This result correlates with the intensive angiogenesis starts the formation of a dense capillary network in the Canalicular phase.



\section {Conclusions}
In this work, we introduced SUPER-IVIM-DC, a DNN approach for the estimation of the IVIM model parameters from DWI data acquired with limited number of b-values. We demonstrated the added-value of SUPER-IVIM-DC over previously proposed DNN-based approach using numerical simulations and healthy volunteer study. Further, we show the potential clinical impact of SUPER-IVIM-DC in assessing fetal lung maturity from DWI data acquired with limited number of b-values.
Thus, SUPER-IVIM-DC has the potential to reduce the overall acquisition times required for IVIM analysis of DWI data in various clinical applications.
Specifically, SUPER-IVIM-DC has the potential to serve as a non-invasive prenatal diagnostic tool to study lung development and to quantify fetal growth. In turn, these data could have clinical relevance as benchmark values to distinguish normal fetuses from pathological fetuses with abnormal lung development.

\bibliographystyle{splncs04}
\bibliography{refs}
\end{sloppypar}

\end{document}